\documentclass[a4paper,conference]{llncs}
\usepackage[latin9]{inputenc}
\PassOptionsToPackage{vlined, ruled}{algorithm2e}
\usepackage{color}
\usepackage{algorithm2e}
\usepackage{graphicx}
\PassOptionsToPackage{normalem}{ulem}
\usepackage{ulem}
\usepackage[unicode=true,pdfusetitle,
 bookmarks=true,bookmarksnumbered=false,bookmarksopen=false,
 breaklinks=false,pdfborder={0 0 1},backref=false,colorlinks=true]
 {hyperref}

\makeatletter

\pdfpageheight\paperheight
\pdfpagewidth\paperwidth

\providecommand{\tabularnewline}{\\}

\usepackage{amsmath}
\usepackage{amssymb}
\usepackage{amsfonts}
\DeclareMathOperator*{\argmax}{argmax}
\DeclareMathOperator*{\argmin}{argmin}

\LinesNumbered

\makeatother

\begin{document}
\title{Efficient Classification with Counterfactual Reasoning and Active
Learning}
\author{Azhar Mohammed, Dang Nguyen, Bao Duong, Thin Nguyen}
\institute{Applied Artificial Intelligence Institute (A\textsuperscript{2}I\textsuperscript{2}),
Deakin University, Geelong, Australia\\
\emph{\{mohammedaz, d.nguyen, duongng, thin.nguyen\}@deakin.edu.au}}
\maketitle
\begin{abstract}
Data augmentation is one of the most successful techniques to improve
the classification accuracy of machine learning models in computer
vision. However, applying data augmentation to \textit{tabular data}
is a challenging problem since it is hard to generate synthetic samples
with labels. In this paper, we propose an efficient classifier with
a novel data augmentation technique for tabular data. Our method called
\textbf{CCRAL} combines causal reasoning to learn counterfactual samples
for the original training samples and active learning to select useful
counterfactual samples based on a \textit{region of uncertainty.}
By doing this, our method can maximize our model's generalization
on the unseen testing data. We validate our method analytically, and
compare with the standard baselines. Our experimental results highlight
that \textbf{CCRAL} achieves significantly better performance than
those of the baselines across several real-world tabular datasets
in terms of accuracy and AUC. Data and source code are available at:
\url{https://github.com/nphdang/CCRAL}.

\end{abstract}


\section{Introduction\label{sec:Introduction}}

Recently, machine learning has become one of the most successful tools
for supporting decisions, and it has been applied widely to many real-world
applications including face recognition \cite{sharif2016accessorize},
security systems \cite{apruzzese2019addressing}, disease detection
\cite{kumar2016dermatological}, or recommended systems \cite{portugal2018use}.
Two core components of a machine learning tool are the algorithm and
the data. The algorithm can be classified into two mainstreams, namely
classification and clustering while the data can be in different formats,
e.g. tabular or image.

When dealing with images in computer vision applications, machine
learning models (or classifiers) often leverage \textit{data augmentation}
techniques to improve the classification accuracy \cite{fawaz2018data}.
The main idea is that given an image of `dog', if we rotate or flip
the image, then we still recognize the object in the image as a `dog'.
By doing this geometric transformation, the label of an image is unchanged
but we can obtain different variants of the image, helping the machine
learning classifier to observe more data and improve its generalization.
In addition to geometric transformation, other data augmentation techniques
are mix-up \cite{zhang2017mixup} and cut-mix \cite{walawalkar2020attentive}.

In spite of a great success in computer vision, applying data augmentation
to tabular data is challenging. There are three main reasons. First,
an image is typically invariant to a small modification, e.g. flip,
zoom, or rotation whereas a small change for a record in tabular data
can result in a totally different outcome. All features (i.e. pixels)
in images are i.i.d (independent and identical distributed) whereas
each feature in tabular data (e.g. Sex or Age) has different ranges
of values. Finally, one transformation operator can be applied to
all features in images whereas each feature in tabular data often
requires a relevant transformation operator depending on the type
of the feature (continuous, discrete or categorical).

\textbf{Our method.} We propose an efficient classification method
with a new data augmentation technique for tabular data. Our method
has two main steps. First, we use causal reasoning to learn counterfactual
samples for the original training samples. Each counterfactual sample
is a variant of an original sample whose all feature values are the
same except the intervened feature. Since the counterfactual samples
may have different outcomes from the original ones, we obtain their
labels via a matching method. Second, we augment counterfactual samples
to real samples to create a new training set to train the classifier.
Since not all counterfactual samples are useful, we select the meaningful
ones that potentially improve the classification performance using
an active learning based method. Our active learning is an uncertain-based
approach. It determines samples that are difficult to predict, then
obtains their counterfactual version to enrich the training data.
Using both real and counterfactual samples, our classifier improves
its generalization, resulting in a better accuracy on unseen testing
samples.

\textbf{Our contribution.} To summarize, we make the following contributions.
\begin{enumerate}
\item We propose \textbf{CCRAL} (\textit{\uline{C}}\textit{lassifier
with }\textit{\uline{C}}\textit{ausal }\textit{\uline{R}}\textit{easoning
and }\textit{\uline{A}}\textit{ctive }\textit{\uline{L}}\textit{earning}),
a novel method for classification with data augmentation in tabular
data. To the best of our knowledge, \textbf{CCRAL} is the first method
that combines both causal reasoning and active learning to train a
classifier with synthetic samples in tabular data.
\item We develop an efficient framework to generate synthetic data. It consists
of two steps: (1) it creates counterfactual samples via sample matching
and (2) it selects useful counterfactual samples via active learning.
\item We demonstrate the benefits of our method on five real-world tabular
datasets, where our method is significantly better than the standard
classifier in both accuracy and AUC measures.
\end{enumerate}
The rest of the paper is organized as follows. In Section \ref{sec:Related-Works},
we briefly outline the fundamentals of data augmentation methods,
the generation of counterfactual data, and active learning in the
literature. We describe our proposed framework \textbf{CCRAL} with
an algorithm and illustrate the \textit{region of uncertainty} in
Section \ref{sec:Framework}. Our experimental settings, datasets,
results are presented in Section \ref{sec:Experiments}, where we
evaluate the performance of \textbf{CCRAL} and compare it with two
existing methods. Finally, we conclude our work in Section \ref{sec:Conclusion}.

\section{Related Works\label{sec:Related-Works}}

\textbf{Data Augmentation}: It is a process of augmenting newly generated
data to the existing training set for improving the model's robustness.
It can be performed by a minor alteration to the existing data. For
example, in computer vision data augmentation is used to enhance deep
learning models by \textit{flipping, color spacing, injecting noise},\textit{
random erasing} to reduce the bias in the classifier to favor more
frequently presented training examples \cite{hernandez2018data,devries2017improved}.
It can also be performed by generating synthetic data to act as a
regularizer and reduce over-fitting while training machine learning
models \cite{shorten2019survey}. Some algorithms such as \textit{data
wrapping,} SMOTE and MaxUp modify real-world examples to create augmented
datasets \cite{baird1992document,chawla2002smote,gong2021maxup}.
However, these methods are exclusively useful for either specific
kinds of data. For example, image recognition dataset or to improve
the performance of a particular algorithm like AGCN \cite{walawalkar2020attentive}.

\textbf{Counterfactual Augmented Data (CAD)}: Another popular method
is to augment data is by using counterfactual reasoning to improve
the generalization of the model. CAD can be generated by using existing
machine learning algorithms by matching closely related samples within
the training set, for example, POLYJUICE to generate text and\textit{
counterfactual image generation} for generating images by using generative
adversarial networks \cite{wu2021polyjuice,neal2018open}. Generating
diverse sets of realistic counterfactuals has proven to improve the
model's training efficiency and overall results \cite{mothilal2020explaining}.
For example, in classification problems, the models trained on CAD
were not sensitive to spurious features unlike modified data \cite{kaushik2019learning,chang2021towards}.
While, in discrimination and fairness literature counterfactual data
substitution and CAD helped to mitigate gender bias by replacing duplicate
text and handling conditional discrimination respectively \cite{maudslay2019s,vzliobaite2011handling}.
However, counterfactually augmented data does not always generalize
better than unaugmented datasets of the same size and may also hurt
the model's robustness \cite{huang2020counterfactually}. There is
a significant gap to explore on the quantity and quality of counterfactual
data needed to be augmented on original dataset by an effective learning
process such that, the model generalizes better and is robust across
various environments.

\textbf{Active learning: }It is a process that learns by an interaction
between oracle and learner agent, it resolves the problem of costly
data labeling in the learning process to improve the obtained model
by making it efficient \cite{cohn1996active,nissim2016improving}.
It can also be implemented on existing classification and predictive
algorithms to optimize a model's performance when compared with state-of-the-art
methods \cite{collet2014active}. For example, in classification problems,
logistic regression yielded remarkably better results by implementing
the simplest suggested active learning method \cite{lewis1994sequential,settles_12_active,yang2018benchmark}.
There are lots of effective approaches such as margin-based methods
\cite{ducoffe2018adversarial} and uncertainty sampling-based methods
to optimize this process \cite{gal2017deep,settles2007multiple}.
By using the uncertainty sampling-based learning process we can measure
how certain a probabilistic classifier's prediction is and, obtain
counterfactual versions of uncertain samples from the \textit{region
of uncertainty} to improve the model's transportability and robustness.

\section{Framework\label{sec:Framework}}

\subsection{Problem definition\label{subsec:Problem-definition}}

Let $f(x)$ be a classifier and $\mathcal{D}=\{x_{i},y_{i}\}_{i=1}^{N}$
be a dataset. Each $y_{i}\in\{0,1\}$ is a binary \textit{true label}.
Given a sample $x_{i}\in{\cal D}$, $f(x_{i})$ provides a probability
(called \textit{predicted score}) that $x_{i}$ belongs to label 1
(i.e. $f(x_{i})=P(y_{i}=1\mid x_{i})$ and $f(x_{i})\in[0,1]$). We
denote the \textit{predicted label} of $x_{i}$ as $\hat{y}_{i}\in\{0,1\}$,
where $\hat{y}_{i}$ is the rounding of $f(x_{i})$ (i.e. $\hat{y}_{i}=1$
if $f(x_{i})\geq0.5$, otherwise $\hat{y}_{i}=0$).
\begin{definition}
(\textbf{Accuracy}). We define accuracy as $P(\hat{y}=y)$, which
means the percentage of samples in $\mathcal{D}$ predicted correctly
by $f(x)$.
\end{definition}

\textbf{Problem statement.} Given a training set $\mathcal{D}_{tr}=\{x_{i},y_{i}\}_{i=1}^{N}$
and a \textit{hold-out} test set ${\cal D}_{te}=\{x_{i},y_{i}\}_{i=1}^{M}$,
our goal is to learn a classifier $f(x)$ using ${\cal D}_{tr}$ such
that $f(x)$ maximizes its accuracy on $\mathcal{D}_{te}$. This is
the traditional classification problem in machine learning \cite{bishop_06_pattern}.

\subsection{Proposed method CCRAL}

A typical way to solve the above problem is to train the classifier
$f(x)$ using the available samples in the training set ${\cal D}_{tr}$,
which tries to minimize a loss function measuring the difference between
the true labels $y$ and the predicted labels $\hat{y}$. Although
this approach is straightforward, it often does not achieve good results.

Our method to solve the classification problem described in Section
\ref{subsec:Problem-definition} is novel. Our main idea is that instead
of using only training samples in ${\cal D}_{tr}$, we try to obtain
more training samples, which is very helpful in improving the generalization
of the classifier. When the classifier observes more training samples,
it is more robust and its classification accuracy is often improved
on \textit{unseen} test samples. This process is often called \textit{data
augmentation}, which has become the state-of-the-art method to improve
the performance of deep learning models in computer vision \cite{shorten2019survey}.

Our approach, called \textit{\uline{C}}\textit{lassifier with }\textit{\uline{C}}\textit{ausal
}\textit{\uline{R}}\textit{easoning and }\textit{\uline{A}}\textit{ctive
}\textit{\uline{L}}\textit{earning} (\textbf{CCRAL}), has two main
steps: (1) learning counterfactual samples using causal reasoning
and (2) training a classifier with both real and counterfactual samples
using active learning.

\subsubsection{Learning counterfactual samples.\label{subsec:Learning-counterfactual-samples.}}

We are dealing with the classification task on \textit{tabular data}.
Typically, a tabular dataset includes a mix of different types of
features. They can be continuous, binary, or categorical features.
Following the standard approach in causal reasoning \cite{wang2015visual},
given the training set ${\cal D}_{tr}$ we select one binary feature
$T$ as the \textit{treatment feature}. For example, the treatment
feature can be Sex=''male/female'' or Marital\_Status=''single/married''.

After determining the treatment feature $T$, we can obtain the counterfactual
of any sample $x\in{\cal D}_{tr}$. Given a sample $x_{i}$, assume
that its treatment feature has value $0$ (i.e. $T_{i}=0$), we then
change the value of the treatment feature to 1. By doing this way,
we now have the counterfactual sample $\bar{x}_{i}$ of $x_{i}$,
which is the same as $x_{i}$ except that the treatment feature of
$\bar{x}_{i}$ has value 1 instead of 0.

Since $\bar{x}_{i}$ is not a real sample, we do not have its label.
To find the label $\bar{y}_{i}$ of $\bar{x}_{i}$, we use the sample
matching approach that computes the distance between $\bar{x}_{i}$
and other samples $x'\in{\cal D}_{tr}$, and uses the label of the
nearest sample as the label of $\bar{x}_{i}$ \cite{bottou2013counterfactual}.
The formulation is retrieved the label of $\bar{x}_{i}$ is as follows:
\begin{equation}
\bar{y}_{i}=y(\argmin_{x'\in{\cal D}_{tr}}d(\bar{x}_{i},x')),\label{eq:distance}
\end{equation}
where $d(\bar{x}_{i},x')$ is the function computing the distance
between the counterfactual sample $\bar{x}_{i}$ and a sample $x'\in{\cal D}_{tr}$.
Any distance can be used, for example, Euclidean, cosine, or Manhattan
distances. In our case, we use the Euclidean distance. The function
$\argmin_{x'\in{\cal D}_{tr}}d(\bar{x}_{i},x')$ returns the sample
that is nearest to $\bar{x}_{i}$, and $y(x_{i})$ is the function
that returns the label of an sample $x_{i}\in{\cal D}_{tr}$.

\subsubsection{Training classifier with real and counterfactual samples.\label{subsec:Training-classifier-with}}

Using Equation (\ref{eq:distance}), we can generate the counterfactual
version of any sample $x\in{\cal D}_{tr}$. The next question is how
to use these counterfactual samples to improve the classification.
Should we create the counterfactual counterpart for each sample, and
augment them to the original training data to train the classifier?
Using all counterfactual samples might not be a good solution. First,
these counterfactual samples are unreal samples, they might add noises
to the training data. Second, the quality of the labels of the counterfactual
samples depend on how we compute the distance in Equation (\ref{eq:distance}).
Finally, in some cases, if there were not very similar samples with
the counterfactual sample $\bar{x}_{i}$, then the label $\bar{y}_{i}$
would be random.

To overcome the three above challenges when using the counterfactual
samples as training data, we propose an \textit{active learning} based
method. We first train a classifier $f(x)$ using samples $x_{i}$
in the training data ${\cal D}_{tr}$. Once we have learned the classifier
$f(x)$, we use it to predict the score for each sample $x_{i}\in{\cal D}_{tr}$.

Since the classifier $f(x)$ has been trained with ${\cal D}_{tr}$,
$f(x)$ predicts confidently the labels for most of the samples in
${\cal D}_{tr}$, where their predicted scores are close to 0 or 1.
However, some samples are difficult to predict their outcomes, where
their scores are close to the decision boundary (i.e. their scores
are close to 0.5). We call these samples are \textit{uncertain samples}.

To determine which samples are uncertain, we define an \textit{uncertain
region} as follows:
\begin{equation}
0.5-\alpha\leq f(x)\leq0.5+\alpha,\label{eq:uncertain_region}
\end{equation}
where $\alpha$ is the \textit{region margin}, $0.5-\alpha$ is the
lower region margin, and $0.5+\alpha$ is the upper region margin.

\begin{figure}
\begin{centering}
\includegraphics[scale=0.65]{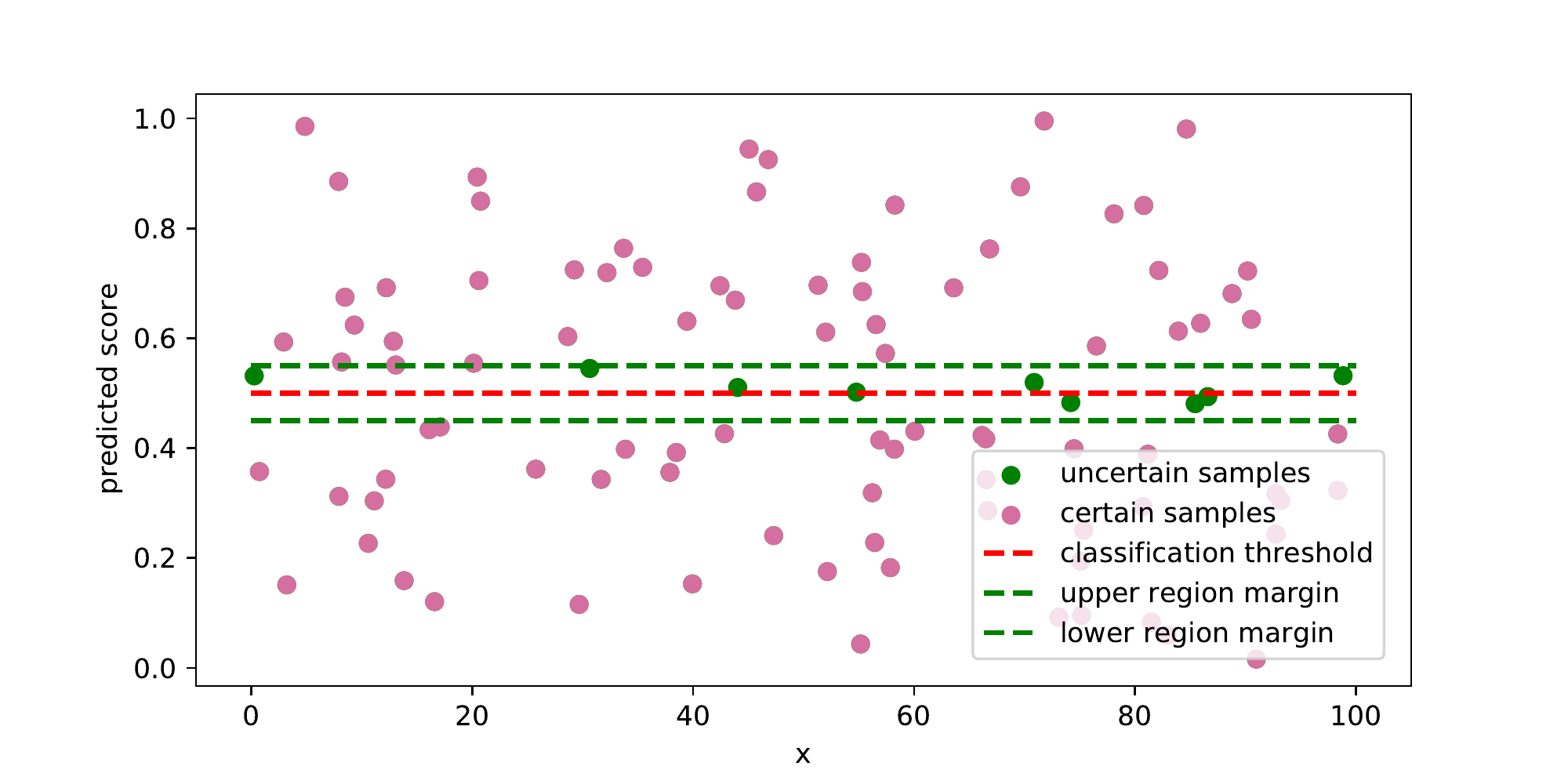}
\par\end{centering}
\caption{\label{fig:Illustration-of-uncertain-samples}Illustration of uncertain
region and uncertain samples. Uncertain samples are indicated by green
circles while the uncertain region is formed by two dashed green lines,
the upper region margin and the lower region margin.}
\end{figure}

From Equation (\ref{eq:uncertain_region}), if any training sample
$x_{i}$ whose predicted score $f(x_{i})$ is in the uncertain region,
then it will be the uncertain sample. Figure~\ref{fig:Illustration-of-uncertain-samples}
illustrates the uncertain region and the uncertain samples.

Since the classifier $f(x)$ is very confused about the label of uncertain
samples. It could be useful if we used their counterfactual version
for the training process. Let ${\cal U}=\{x_{1},x_{2},...,x_{n}\}$
be the set of uncertain samples. Following the process in Section
\ref{subsec:Learning-counterfactual-samples.}, we learn counterfactual
version for each sample $x_{i}\in{\cal U}$. We then augment these
counterfactual samples $\bar{{\cal U}}=\{\bar{x}_{1},\bar{x}_{2},...,\bar{x}_{n}\}$
to the original training set ${\cal D}_{tr}$ i.e. we have the new
training set ${\cal D}'_{tr}={\cal D}_{tr}\cup\bar{{\cal U}}$. Finally,
we train the classifier $f(x)$ again with the new training set ${\cal D}'_{tr}$.

Since the region margin $\alpha$ has values being in the range of
$[0,0.5]$, we use a grid search (or Bayesian optimization \cite{nguyen2020bayesian})
to find the $\alpha$ that derives the best classifier $f(x)$ measured
on a validation set ${\cal D}_{va}$. In particular, at each search
iteration, we expand the uncertain region by increasing the value
of $\alpha$, and obtain more uncertain samples. We then find the
counterfactual counterparts of these uncertain samples. Finally, we
train the classifier $f(x)$ with real training samples along with
the counterfactual samples and measure its accuracy on a validation
set. The final classifier is the classifier whose accuracy is highest
on the validation set, and this final classifier will be evaluated
on the hold-out test set.

Algorithm \ref{alg:The-proposed-CCRAL} summarizes our method \textbf{CCRAL}.

\begin{algorithm}
\KwIn{$\mathcal{D}=\{x_{i},y_{i}\}_{i=1}^{N}$: training set, $K$:
\# of iterations}

{

split ${\cal D}$ into a (smaller) training set ${\cal D}_{tr}$ and
a validate set ${\cal D}_{va}$\;

define a grid of margins $[\alpha_{1},\alpha_{2},...,\alpha_{K}]$\;

train a classifier $f(x)$ \textit{\emph{on }}$\mathcal{D}_{tr}$\;

select a binary feature $T$ as the treatment feature\;

\For{each sample $x_{i}\in{\cal D}_{tr}$ }{

generate its counterfactual sample $\bar{x}_{i}$ by changing the
value of the treatment feature of $x_{i}$\;

compute its counterfactual label $\bar{y}_{i}=y(\argmin_{x'\in{\cal D}_{tr}}d(\bar{x}_{i},x'))$
(see Equation (\ref{eq:distance}))\;

}

use $f(x)$ to predict a score $f(x_{i})$ for each sample $x_{i}\in{\cal D}_{tr}$\;

\For{$k=1,2,...,K$ }{

find ${\cal U}_{k}=\{x_{1},x_{2},...,x_{n}\}$, where $x_{i}$ is
an uncertain sample i.e. $0.5-\alpha_{k}\leq f(x_{i})\leq0.5+\alpha_{k}$
(see Equation (\ref{eq:uncertain_region}))\;

generate new training data ${\cal D}_{tr}^{k}={\cal D}_{tr}\cup\bar{{\cal U}}^{k}$
where $\bar{{\cal U}}^{k}=\{\bar{x}_{1},\bar{x}_{2},...,\bar{x}_{n}\}$
is the counterfactual of ${\cal U}^{k}$\;

train $f_{k}(x)$ on ${\cal D}{}_{tr}^{k}$\;

evaluate accuracy $acc_{k}$ of $f_{k}(x)$ on ${\cal D}_{va}$\;

}

return the best classifier $f_{k^{*}}(x)$, where $k^{*}=\argmax_{k}acc_{k}$\;

}

\caption{\label{alg:The-proposed-CCRAL}The proposed \textbf{CCRAL} algorithm.}

\end{algorithm}

\section{Experiments and Discussions\label{sec:Experiments}}

We conduct extensive experiments on five real-world tabular datasets
to evaluate the classification performance (accuracy and AUC) of our
method \textbf{CCRAL}, comparing it with two strong baselines.

\subsection{Datasets}

To create an environment for comprehending counterfactual reasoning
involved in our method \textbf{CCRAL}, we choose five real-world tabular
datasets that have at least one binary feature that intrigues one's
causal thinking. These datasets were often used to evaluate fairness-aware
and causal inference machine learning algorithms \cite{friedler2019comparative,zafar2017parity,nguyen2021fairness}.

Table~\ref{tab:Characteristics-of-datasets} shows characteristics
of each dataset along with the selected treatment feature and the
respective outcome.

\begin{table}
\caption{\label{tab:Characteristics-of-datasets}Characteristics of five tabular
datasets. We denote $N$: the number of samples, $M$: the number
of features, $T$: the treatment feature, and $y$: the class feature.}

\centering{}%
\begin{tabular}{|l|r|r|c|c|c|c|c|c|}
\hline 
\textbf{Dataset} & \textbf{$N$} & \textbf{$M$} & $T$ & \textbf{$T=1$} & \textbf{$T=0$} & \textbf{$y$} & \textbf{$y=1$} & \textbf{$y=0$}\tabularnewline
\hline 
\hline 
\textit{german} & 1,000 & 20 & Sex & ``male'' & ``female'' & Credit & ``good'' & ``bad''\tabularnewline
\hline 
\textit{bank} & 4,521 & 14 & Marriage & ``married'' & ``single'' & Subscription & ``yes'' & ``no''\tabularnewline
\hline 
\textit{twins} & 4,821 & 52 & Weight & ``heavier'' & ``lighter'' & Mortality & ``alive'' & ``death''\tabularnewline
\hline 
\textit{compas} & 4,010 & 10 & Sex & ``male'' & ``female'' & Rearrested & ``no'' & ``yes''\tabularnewline
\hline 
\textit{adult} & 30,162 & 13 & Sex & ``male'' & ``female'' & Income & ``\textgreater 50K'' & ``\textless 50K''\tabularnewline
\hline 
\end{tabular}
\end{table}
\textit{german}: this dataset describes each individual's credit score
whether she/he has a good or bad credit score \cite{dua_graff_19_uci}.
It has 1,000 samples and 20 features. We use \textit{Sex} as the treatment
feature.

\textit{bank}: this dataset is about direct marketing campaigns of
individuals for term deposit subscriptions. The outcome of this data
is whether a person is subscribed or not depending upon the marketing
and duration campaigned. \textit{Marriage} is the treatment feature
in this dataset.

\textit{twins}: this dataset consists of around 5,000 records of twin's
birth collected during the period of 1989-1991 in the U.S. \cite{almond2005costs}.
It is a popular benchmark dataset in causality researches \cite{louizos2017causal}.
The outcome corresponds to the mortality of each twin's during the
first year of birth. We choose twins of the same gender to replicate
the counterfactual. The treatment feature is the twin's\textit{ weight}.

\textit{compas}: this dataset includes a collection of data in Broward
country, Florida about the use of the COMPAS risk assessment tool
and has the data regarding felonies and charges on the degree of the
arrest \cite{angwin2016machine}. This dataset has the treatment feature
\textit{Sex} with an outcome of getting rearrested within two years.

\textit{adult}: this dataset is the collection of individual data
of their income recorded during the 1994 U.S census \cite{kamiran2012decision}.
The outcome is a person's income. If the income is greater than \$50K,
then it is labeled as ``1''. Otherwise it is ``0''. This dataset
has 30,162 samples and 13 features. We select \textit{Sex} as the
treatment feature.

\subsection{Baselines and evaluation}

We compare our method \textbf{CCARL} with two strong baselines.
\begin{enumerate}
\item \textbf{Standard}: this method uses available training samples to
train a classifier.
\item \textbf{Counterfactual}: this method uses the counterfactual samples
of all original training samples in the training process. In other
words, it fixes $\alpha=0.5$ in Equation (\ref{eq:uncertain_region}).
\end{enumerate}
For a fair comparison, we measure the accuracy and AUC of each method
on the same hold-out test set. We also use the same classifier for
all methods, namely the Support Vector Machine (SVM) with the linear
kernel and $C=1$ for the regularization. Note that other machine
learning classifiers can be used with our method. We use the default
search range $[0,0.5]$ for $\alpha$, and set the number of iterations
$K=10$. We evaluate methods on each dataset in five times with different
train-test data splits, and report the averaged accuracy and AUC.

\subsection{Results}

\begin{figure}
\centering{}\includegraphics[scale=0.35]{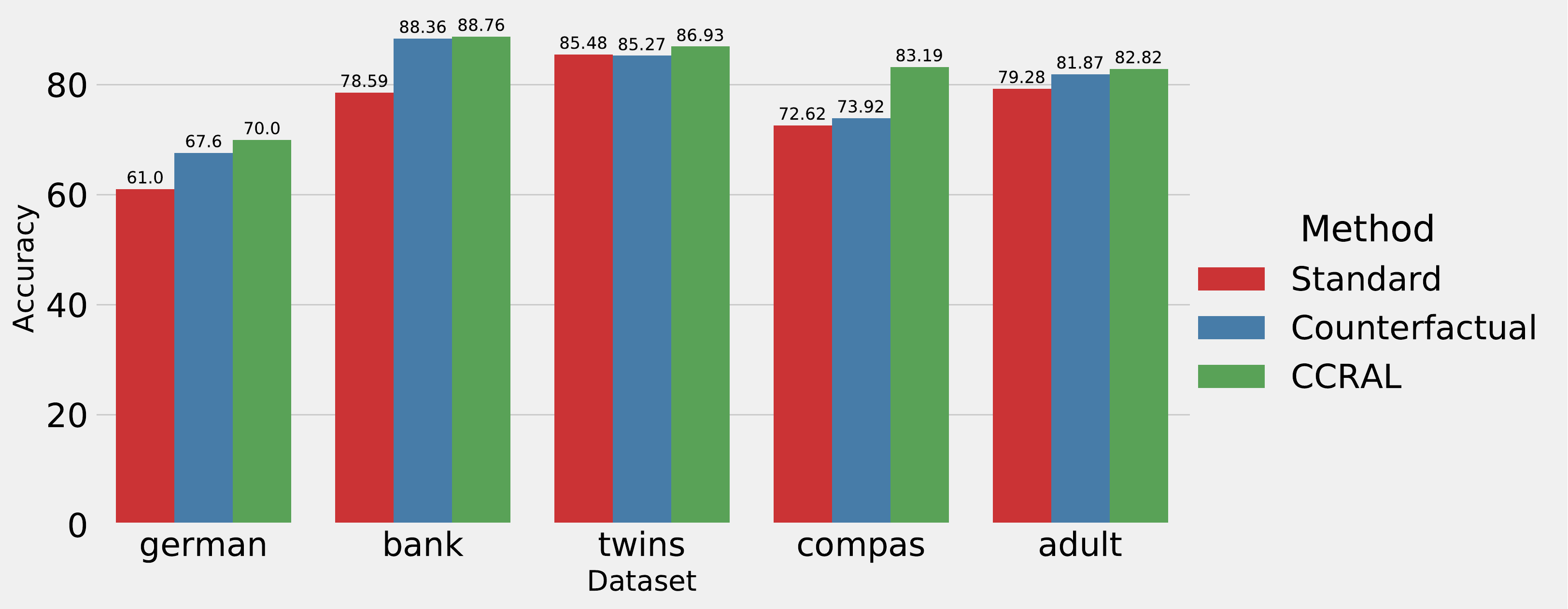}\caption{\label{fig:Classification-accuracy}The averaged classification accuracy
of two baselines Standard, Counterfactual, and our method \textbf{CCRAL}
on each dataset.}
\end{figure}

Figure \ref{fig:Classification-accuracy} shows the accuracy of each
method on five datasets. It can be seen that our method \textbf{CCRAL}
is much better than the standard classifier on all datasets. On \textit{german}
(a small dataset), Standard achieves only 61.0\% whereas \textbf{CCRAL}
achieves 70.0\%, resulting in 9\% better. On \textit{adult} (a very
large dataset), the accuracy of Standard is 79.28\% compared to 82.82\%
of our \textbf{CCRAL}. On this dataset, our method achieves around
3\% gains over the standard classifier.

Compared to the Counterfactual method, \textbf{CCRAL} is comparable
on three datasets \textit{bank}, \textit{twins}, and \textit{adult}
while it is much better on two datasets \textit{german} and \textit{compas}.
This shows that using all counterfactual samples in the training process
was not a good solution since they might add noise to the training
data, as we discussed in Section \ref{subsec:Training-classifier-with}.
Our method which uses active learning to select useful counterfactual
samples is a more efficient approach to train the classifier.

\begin{figure}
\begin{centering}
\includegraphics[scale=0.35]{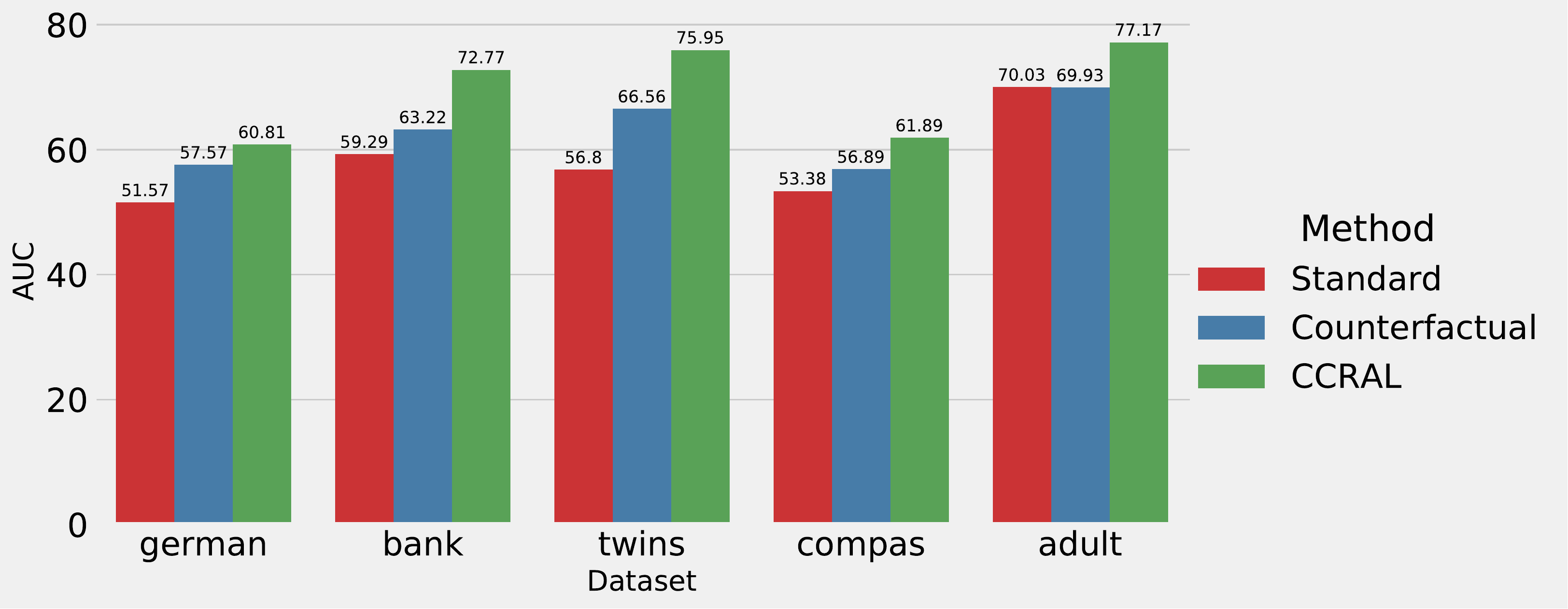}
\par\end{centering}
\caption{\label{fig:Classification-AUC}The averaged AUC of two baselines Standard,
Counterfactual, and our method \textbf{CCRAL} on each dataset.}
\end{figure}

We also report the AUC of each method in Figure \ref{fig:Classification-AUC}.
Our \textbf{CCRAL} is the best method, where it significantly outperforms
two baselines Standard and Counterfactual. \textbf{CCRAL} always outperforms
the standard classifier by a large margin across all datasets. Compared
to Counterfactual, our method shows a great improvement, where it
achieves 3-9\% gains over Counterfactual. Again, this suggests that
using active learning to select useful counterfactual samples is a
much better strategy than using all counterfactual samples for training
the classifier.

\section{Conclusion\label{sec:Conclusion}}

In this paper, we have introduced an efficient classifier (named \textbf{CCRAL})
with a novel data augmentation technique for tabular datasets. We
generate counterfactual data by flipping the binary value of the treatment
feature of original training samples, and obtain their labels by using
a matching method. We use active learning to select useful counterfactual
samples based on a \textit{region of uncertainty} depending on the
predicted scores of the original training samples. We augment selected
counterfactual samples to the set of original training samples to
train the classifier. We demonstrate the efficacy of \textbf{CCRAL}
on five standard real-world tabular datasets. The obtained results
show that \textbf{CCRAL} generalizes better and is more robust towards
unseen testing samples, where it significantly outperforms other methods.
Our approach can be conceptually extended to other types of data such
as sequences \cite{nguyen2018sqn2vec} and graphs \cite{nguyen2018learning}.

\section*{Acknowledgment}

This research is partly supported by NHMRC Ideas Grant GNT2002234.

\bibliographystyle{splncs04}
\bibliography{reference}

\end{document}